\documentclass[conference]{IEEEtran}
\IEEEoverridecommandlockouts
\usepackage{cite}
\usepackage{amsmath,amssymb,amsfonts,commath}
\usepackage{algorithmic}
\usepackage{graphicx}
\usepackage{textcomp}
\usepackage{xcolor}
\usepackage{siunitx}

\def\BibTeX{{\rm B\kern-.05em{\sc i\kern-.025em b}\kern-.08em
    T\kern-.1667em\lower.7ex\hbox{E}\kern-.125emX}}
\begin{document}

\title{Data-driven Full-waveform Inversion Surrogate using Conditional Generative Adversarial Networks

\thanks{We thank Petrobras S.A. for providing computational resources and the data used in this publication.}
}

\makeatletter
\newcommand{\linebreakand}{%
  \end{@IEEEauthorhalign}
  \hfill\mbox{}\par
  \mbox{}\hfill\begin{@IEEEauthorhalign}
}
\makeatother

\author{
	\IEEEauthorblockN{
		Marcus Saraiva$^{a,c}$\IEEEauthorrefmark{3},
		Avelino Forechi$^{b}$\IEEEauthorrefmark{2},
		Jorcy de Oliveira Neto$^{c}$\IEEEauthorrefmark{3},
		Antônio DelRey$^{c}$\IEEEauthorrefmark{3},
		Thomas Rauber$^{a}$\IEEEauthorrefmark{1}
	}
	\and
	\IEEEauthorblockA{
		\hspace{1cm}$^{a}$\textit{Departamento de Inform\'atica -- Centro Tecnol\'ogico}\\
		\hspace{1cm}\textit{Universidade Federal do Esp\'irito Santo}\\
		\hspace{1cm}Vit\'oria, Esp\'irito Santo, Brazil\\
		\hspace{1cm}\IEEEauthorrefmark{1}thomas.rauber@ufes.br
	}
	\and
	\IEEEauthorblockA{
		$^{b}$\textit{Coordenadoria de Engenharia Mec\^anica}\hspace{1cm}\\
		\textit{Instituto Federal do Esp\'irito Santo}\hspace{1cm}\\
		Aracruz, Esp\'irito Santo, Brazil\hspace{1cm}\\
		\IEEEauthorrefmark{2}avelino.forechi@ifes.edu.br\hspace{1cm}
	}	
	\and
	\IEEEauthorblockA{\centerline{
		$^{c}$\textit{Explora\c c\~ao e Produ\c c\~ao}} \\
		\textit{Petróleo Brasileiro S.A.}\\
		Vit\'oria, Esp\'irito Santo, Brazil\\
		\IEEEauthorrefmark{3}\{mvsaraiva, jorcyneto, delrey\}@petrobras.com.br
	}
}

\maketitle
\begin{abstract}

In the Oil and Gas industry, estimating a subsurface velocity field is an essential step in seismic processing, reservoir characterization, and hydrocarbon volume calculation.
Full-waveform inversion (FWI) velocity modeling is an iterative advanced technique that provides an accurate and detailed velocity field model, although at a very high computational cost due to the physics-based numerical simulations required at each FWI iteration. 
In this study, we propose a method of generating velocity field models, as detailed as those obtained through FWI, using a conditional generative adversarial network (cGAN) with multiple inputs.
The primary motivation of this approach is to circumvent the extremely high cost of full-waveform inversion velocity modeling.
Real-world data were used to train and test the proposed network architecture, and three evaluation metrics (percent error, structural similarity index measure, and visual analysis) were adopted as quality criteria.
Based on these metrics, the results evaluated upon the test set suggest that the GAN was able to accurately match real FWI generated outputs, enabling it to extract from input data the main geological structures and lateral velocity variations. 
Experimental results indicate that the proposed method, when deployed, has the potential to increase the speed of geophysical reservoir characterization processes, saving on time and computational resources.
 
\end{abstract}

\section{Introduction}

Oil industry applications utilize seismic reflection as a primary indirect method for investigating the properties of subsurface strata. This method uses a controlled source of acoustic waves and receiver arrays (geophones or hydrophones) to record the transit time and amplitude of acoustic waves reflected by contrasting rock type interfaces (Fig. \ref{fig:Seismic_Aquisition}).

\begin{figure}[t]
     \centering
     \includegraphics[width=0.80\hsize]{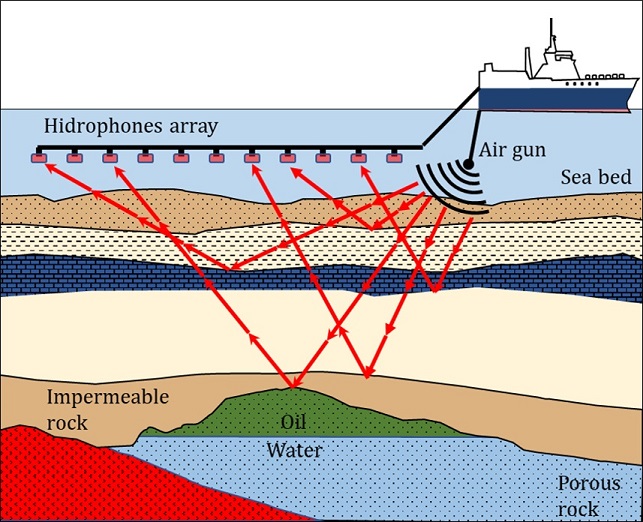}
      \caption{Schematic representation of a marine seismic acquisition.}
     \label{fig:Seismic_Aquisition}
\end{figure}

During seismic acquisition, the same subsurface region is sampled multiple times from different offsets, resulting in data redundancy. The vertical projection of this region on the surface is called the common midpoint (CMP). Sorting the data by CMP, we can then obtain the so-called CMP gathers
(Fig. \ref{fig:Seismic_Stack}).

To obtain an image of the subsurface area, it is then necessary to process the recorded data (transit time and amplitude of the reflected acoustic waves). 
The estimation of an accurate velocity model is an essential step in seismic processing. This model is used in several steps of the seismic processing, including seismic migration, normal moveout (NMO) correction, and time-to-depth conversion.

Seismic migration is a procedure that aims to "migrate" seismic events to their correct position and collapse reflections and diffractions, creating a seismic image of the subsurface\cite{Gazdag_Sguazzero}.
NMO correction adjusts all CMP seismic traces to a theoretical zero offset (source and receiver at the same position)\cite{Yilmaz}. This procedure enables the use of data redundancy to increase signal-to-noise ratio through CMP-corrected gather stacking (Fig. \ref{fig:Seismic_Stack}).
The velocity model is also important for time-to-depth conversion, a necessary step for reservoir characterization and oil volume estimation \cite{schultz1998seismic}.

\begin{figure}[t]
     \centering
     \includegraphics[width=1.\hsize]{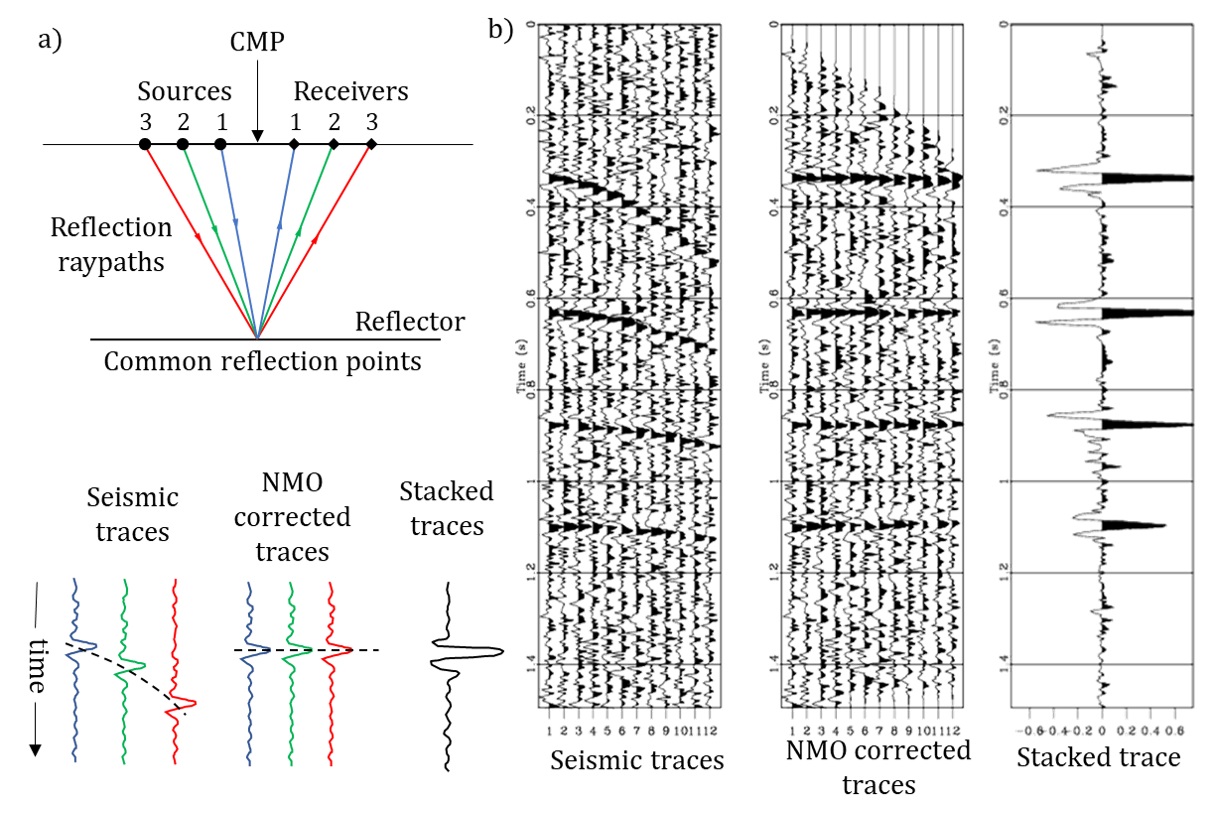}
      \caption{a) Schematic representation of normal moveout correction; b) Process of stacking a synthetic CMP gather after normal moveout correction to improve the signal-to-noise ratio (Adapted from \cite{Liu2009StackingSD}).}
     \label{fig:Seismic_Stack}
\end{figure}

Complex geological settings such as salt domes and folded or faulted regions require detailed velocity models in order to form a satisfactory seismic image. Currently, velocity modeling using full-waveform inversion (FWI) is an advanced technique providing an accurate and detailed velocity model~\cite{Kapoor2012FullWI}.

In this paper, we present a method using an image-to-image translator (pix2pix)\cite{isola2018imagetoimage} based on a conditional generative adversarial network (cGAN) to translate an input (post-stack seismic image, average tomographic velocity, and two-way time grid) into an interval velocity model as detailed as FWI.

\subsection{Seismic data and velocity model}

The primary outcome of 3D seismic processing is a seismic image volume. Seismic volume slices in the acquisition direction are conventionally referred to as inlines, slices perpendicular to the acquisition direction as crosslines, slices perpendicular to the vertical axis as time slices, and a vertical sample vector as a seismic trace (Fig. \ref{fig:Seismic_Volume}).

\begin{figure}[h]
     \centering
     \includegraphics[width=0.80\hsize]{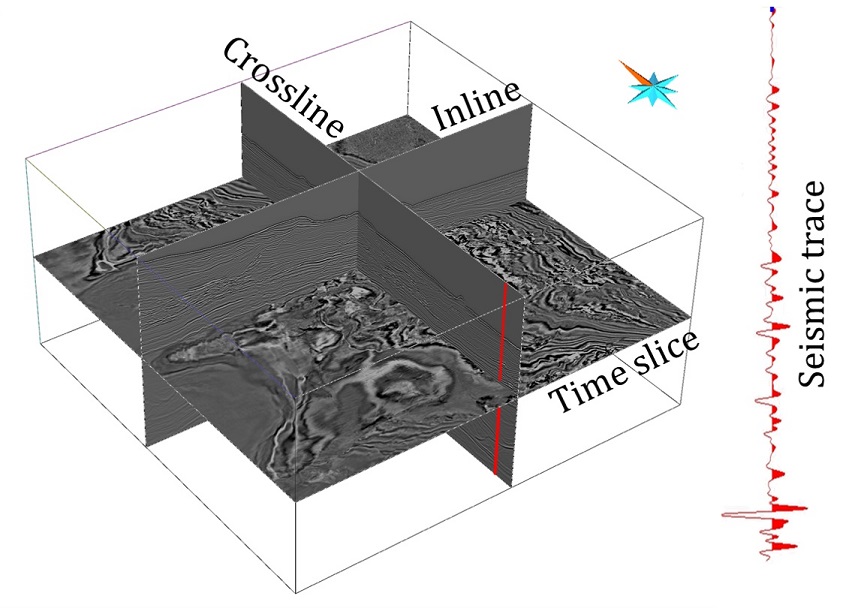}
      \caption{3D representation of seismic image and main slicing directions: inline (parallel to seismic acquisition direction), crossline (perpendicular to seismic acquisition direction) and time slice (perpendicular to vertical axis). The red line shows a seismic trace (vertical sample vector). }
     \label{fig:Seismic_Volume}
\end{figure}

Seismic volume is among the most important inputs to characterizing hydrocarbon reservoirs. After  the data’s correlation with well logs, it is possible to map the main seismic horizons and delimit the top and bottom of hydrocarbon reservoirs.

A typical workflow for seismic velocity estimation consists of three steps, as follows.

\begin{enumerate}

\item\textbf{Normal moveout (NMO) velocity estimation} is performed by manually picking  velocities that best horizontalize hyperbola reflections on CMP gathers (Fig.~\ref{fig:NMO}). This process returns a very smooth average NMO velocity model that can be approximated by a root mean square velocity for small offsets. 

\item\textbf{Ray-based} or \textbf{grid tomography} is an iterative technique using seismic reflection's travel time measurements and associated amplitudes to calculate seismic velocity \cite{Jones2010TutorialVE}. Seismic tomography delivers a low-frequency seismic acoustic velocity model, typically up to  2–3 Hz, and commonly takes the NMO velocity as the first input.

\item\textbf{Full-waveform inversion (FWI)} is a seismic inversion technique, initially proposed by \cite{lailly1983sequence} and \cite{Tarantola1984InversionOS}, using pre-stack seismic data (raw data from seismic acquisition) to output a detailed interval velocity model.

\end{enumerate}
\begin{figure}[b]
 \centering
 \includegraphics[width=1.0\hsize]{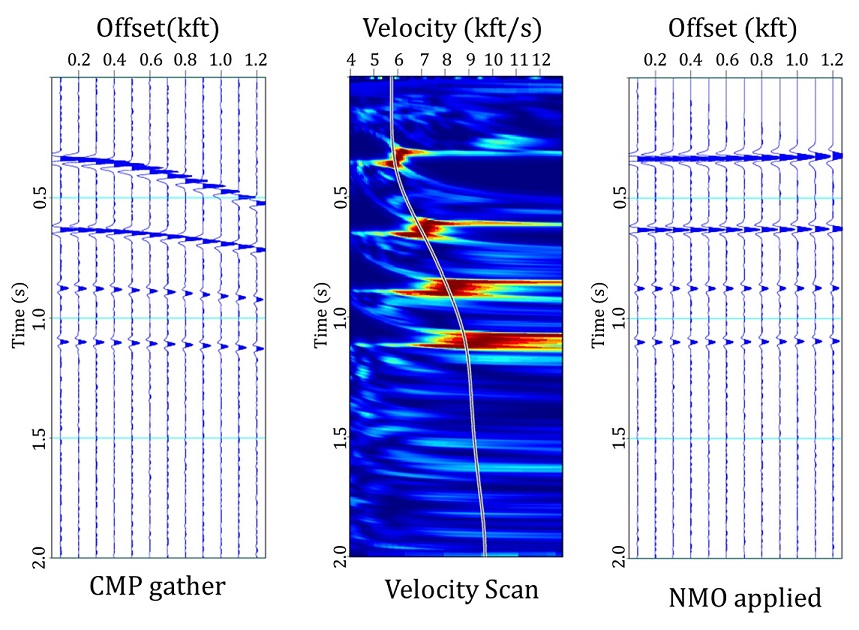}
  \caption{Normal moveout velocity picking. The line passing by the red spots on the central image was manually selected to define the velocity that best horizontalize the reflections at the CMP gather (adapted from \cite{Madagascar}). }
 \label{fig:NMO}
\end{figure}

FWI takes an initial low-resolution velocity model as input to generate synthetic pre-stack seismic data by wave equation numerical modeling. This synthetic model is then compared with real data from seismic acquisition. FWI iteratively updates this velocity model in order to minimize the difference between real and synthetic data (Fig. \ref{fig:FWI}).

\begin{figure*}[t]
 \centering
 \includegraphics[width=1.0\hsize]{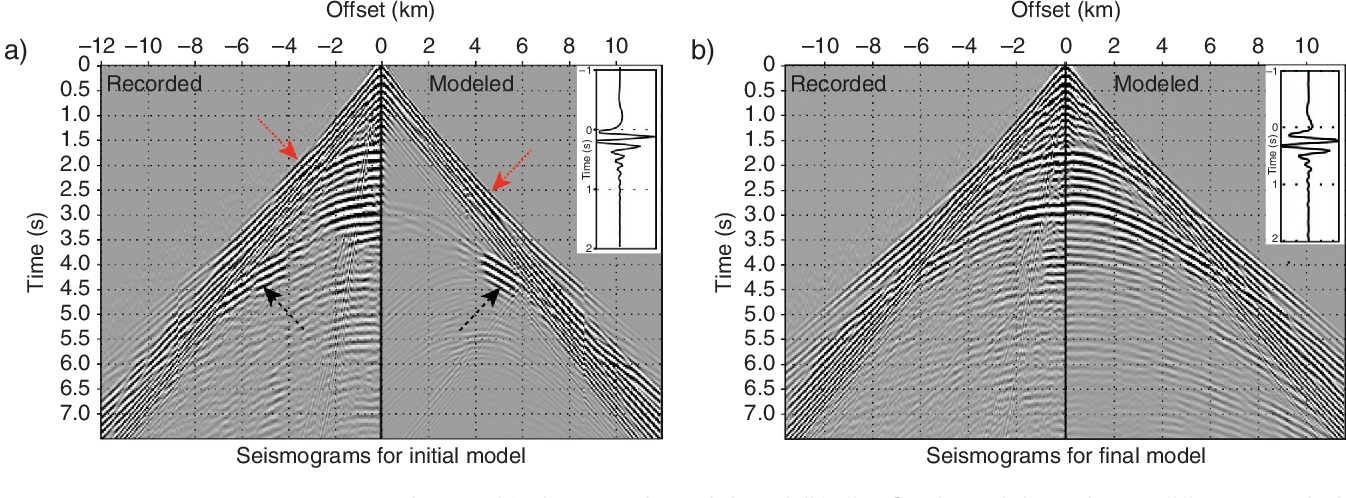}
  \caption{FWI modeling: a) Real seismogram compared with synthetic seismogram generated from the initial velocity model and b) Real seismogram compared with synthetic seismogram generated from the final velocity model (From \cite{virieux2017introduction}). }
 \label{fig:FWI}
\end{figure*}

As FWI is performed with pre-stack seismic data (up to three orders of magnitude greater than the traditional seismic volume), it becomes computationally expensive and time-consuming.  The use of this technique has only become industrially feasible in recent years, with an increase in processing power, although the theory was established in the 1980s \cite{lailly1983sequence}, \cite{Tarantola1984InversionOS}.

Building an FWI velocity model may require from three months up to a year of construction, depending on the seismic acquisition size and geological subsurface complexity. Theoretically, FWI can estimate a high-resolution velocity model using as initial input a very smooth velocity model. In practice, to support the convergence and reduce the number of iterations, it is conventional to take a tomographic velocity model as input.

The FWI velocity model fills the gap between low-frequency tomographic velocity model and the seismic reflection image volume (Fig. \ref{fig:FWIgap}), increasing the useful data bandwidth. 

\subsection{Generative Adversarial Networks (GAN)}

Convolutional neural networks (CNN) are getting more and more popular in geophysical applications, typically in classification and segmentation tasks \cite{Geng2020AutomatedDO}, \cite{Xiong2018SeismicFD}, \cite{Shi2019SaltSegA3},\cite{Wu2019FaultSeg3DUS}, and \cite{Guazzelli2020Efficient3S}. Moreover, recent studies have demonstrated the feasibility of geophysical modeling using CNNs \cite{Ren2020APN}, \cite{Moseley2020SolvingTW}, and \cite{Zhang2019PhysicsguidedCN}.

\begin{figure}[b]
 \centering
 \includegraphics[width=0.80\hsize]{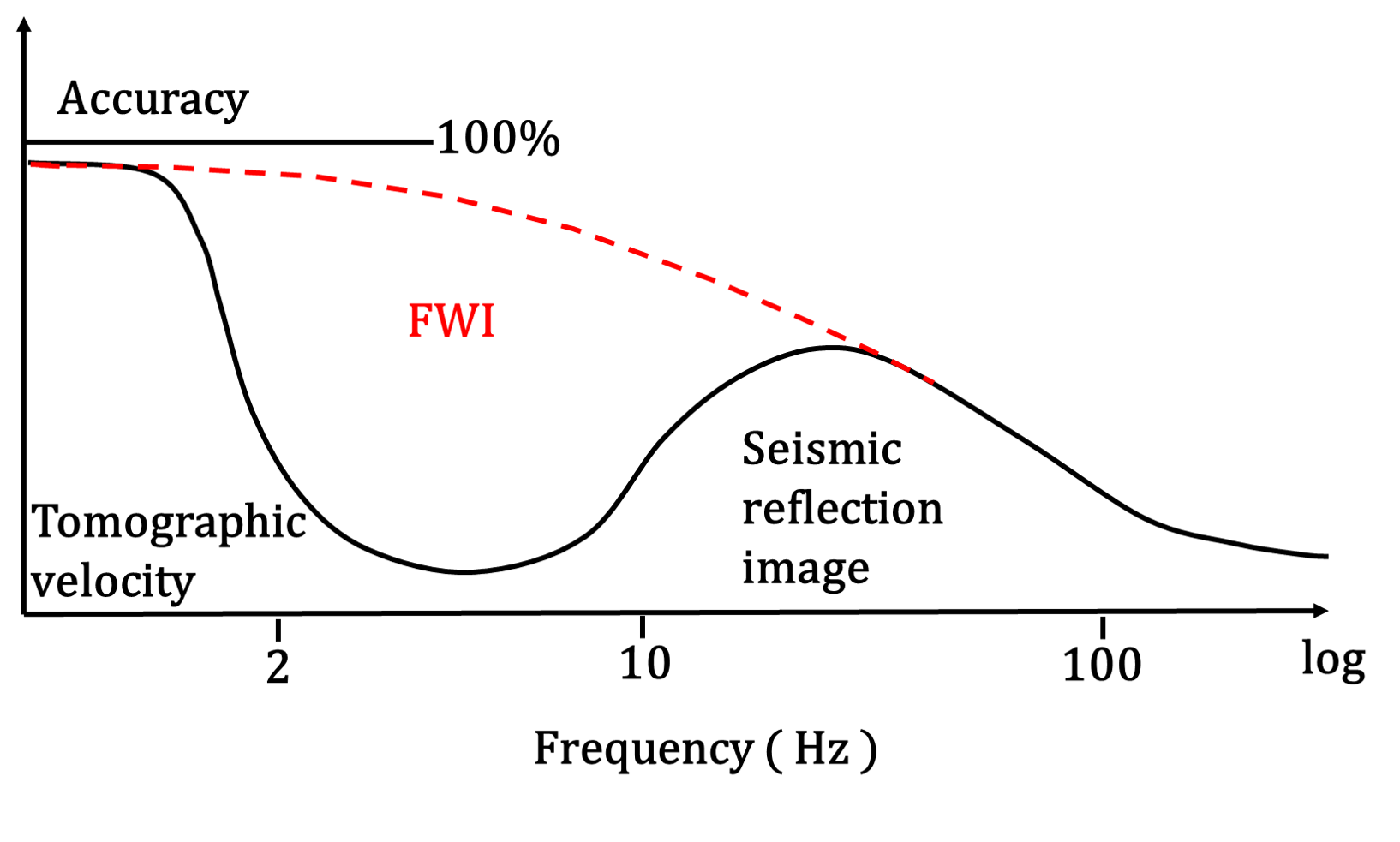}
  \caption{Simplified representation of the information gap between tomographic velocity model and seismic reflection image volume. FWI velocity model fills this gap (Adapted from\cite{Claerbout1985ImagingTE}).}
 \label{fig:FWIgap}
\end{figure}

Among CNN architectures, generative adversarial networks (GAN) go beyond traditional classification and segmentation tasks; GANs stand out for their ability to learn how to mimic data provided to them by capturing their statistical and spatial distributions \cite{Goodfellow2014GenerativeAN}.

GANs are generative models that learn to generalize a rule of how to transform a random noise vector $z$ to output image $y$ ($G: z\rightarrow y$) \cite{Goodfellow2014GenerativeAN}. The pix2pix GAN architecture, proposed by \cite{isola2018imagetoimage} and adopted in this paper, is a conditional GAN (cGAN), which has as its input the sum of the random noise vector $z$ with an image $x$, to find a rule of how to translate this input to image $y$ ($G: \{x,z\}\rightarrow y$).

To accomplish this task, pix2pix uses two networks competing against each other. While one network generates synthesized images (a generator), the other network judges whether these images are valid (a discriminator). Generator G is trained to produce output images that are indistinguishable from the actual images by a discriminator D, which in turn is trained to improve its performance on detecting the generator’s fake images.

In cGAN, the generator output $y$ is connected directly to the discriminator input, together with the image $x$. 
Through the backpropagation process, discriminator classification in real or fake images provides a signal for the generator to update its weights and bias, thus improving performance.
The objective function of a cGAN can be expressed by (\ref{cGAN_equation}).  

\begin{align}
    \mathcal{L}_{\mathrm{cGAN}}(G,D) &= \mathbb{E}_{x,y}\left[\log D(x,y)\right] + \nonumber \\
                 & \mathbb{E}_{x,z}\left[\log (1-D(x,G(x,z))\right],\label{cGAN_equation}
\end{align}

where the generator $G$ tries to minimize this objective function against an adversarial discriminator $D$ that tries to maximize it (\ref{cGAN_equationmax}). 

\begin{align}
     G^*  = \arg\min_G\max_D \mathcal{L}_{\mathrm{cGAN}}(G,D).\label{cGAN_equationmax}
\end{align}

Previous work \cite{Pathak2016ContextEF} shows that L1 regularization is beneficial to this process, helping the generator $G$ to create sharper images (\ref{L1_equation}).

\begin{align}
    \mathcal{L}_{L1}(G) = \mathbb{E}_{x,y,z}\left[\norm{y-G(x,z)}_1\right].\label{L1_equation}
\end{align}

Thus, the final objective function for the pix2pix conditional GAN used in this work is (\ref{full_objective}).

\begin{align}
    G^*  = \arg\min_G\max_D \mathcal{L}_{\mathrm{cGAN}}(G,D) + \lambda \mathcal{L}_{L1}(G).\label{full_objective}
\end{align}

Many image processing problems involve transforming one image into another. These problems are often addressed using algorithms specific to each situation. Conditional GANs are a general solution with good performance on a variety of problems \cite{isola2018imagetoimage}. Fig. \ref{fig:example_pix2pix} shows some generic examples using pix2pix cGAN.

\begin{figure}[b]
 \centering
 \includegraphics[width=1.0\hsize]{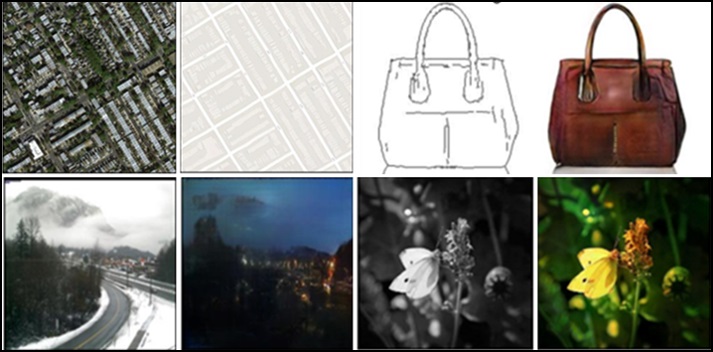}
  \caption{Examples using pix2pix GAN:  aerial photos to maps, edges to photos, day to night, and gray scale to color (Adapted from \cite{isola2018imagetoimage}).}
 \label{fig:example_pix2pix}
\end{figure}

\section{Materials and Methods}

\subsection{Data}
From a marine seismic acquisition of approximately 680~$km^2$  in the Campos Basin, Brazil, we used four seismic volumes:
\begin{itemize}
    \item Post stack seismic image
    \item Average tomographic velocity 
    \item Two-way time (TWT) grid
    \item FWI interval velocity 
\end{itemize}

We loaded the data, originally in seg-y format,  as a 3D matrix in a Python environment using the Segyio library \cite{segyiodocs}. The shape of each matrix was $1451\times3001\times751$, where the $x$ and $y$ samples were spaced $12.5~m$ apart and $z$ samples spaced every $0.004~s$.

Preprocessing involved limiting data variability in two standard deviations to eliminate outliers, partially cropping regions without reflections (sea water thickness), and later data normalization using a global maximum and minimum as parameters (\ref{eq:Norm}).

\begin{align}
     X_{\mathrm{norm}}  = \frac{X-X_{\mathrm{min}}}{X_{\mathrm{max}}-X_{\mathrm{min}}}. \label{eq:Norm}
\end{align}

We divided the data into training and test sets at a ratio of 70\% to 30\%, respectively (Fig. \ref{fig:train_val}). 
For the training set, we randomly sampled 2000 patches (512 × 512). For the test set, we randomly sampled 800 patches (512 × 512), both in the $x$ direction (direction of acquisition).

\begin{figure}[b]
 \centering
 \includegraphics[width=0.9\hsize]{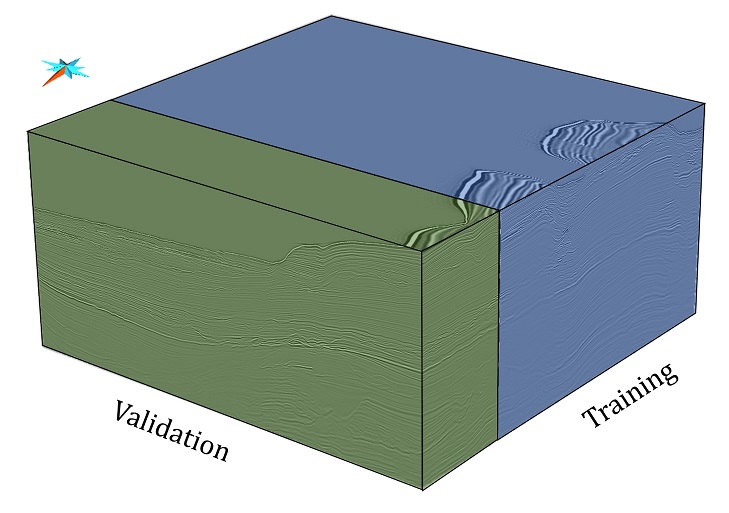}
  \caption{Seismic volume showing the spatial distributions of training and test sets.}
 \label{fig:train_val}
\end{figure}

The post-stack seismic image, average tomographic velocity, and TWT grid were combined to form a three-channel image. The training pair was composed of this image and the FWI interval velocity (Fig. \ref{fig:training_pair}).

\begin{figure}[t]
 \centering
 \includegraphics[width=1.0\hsize]{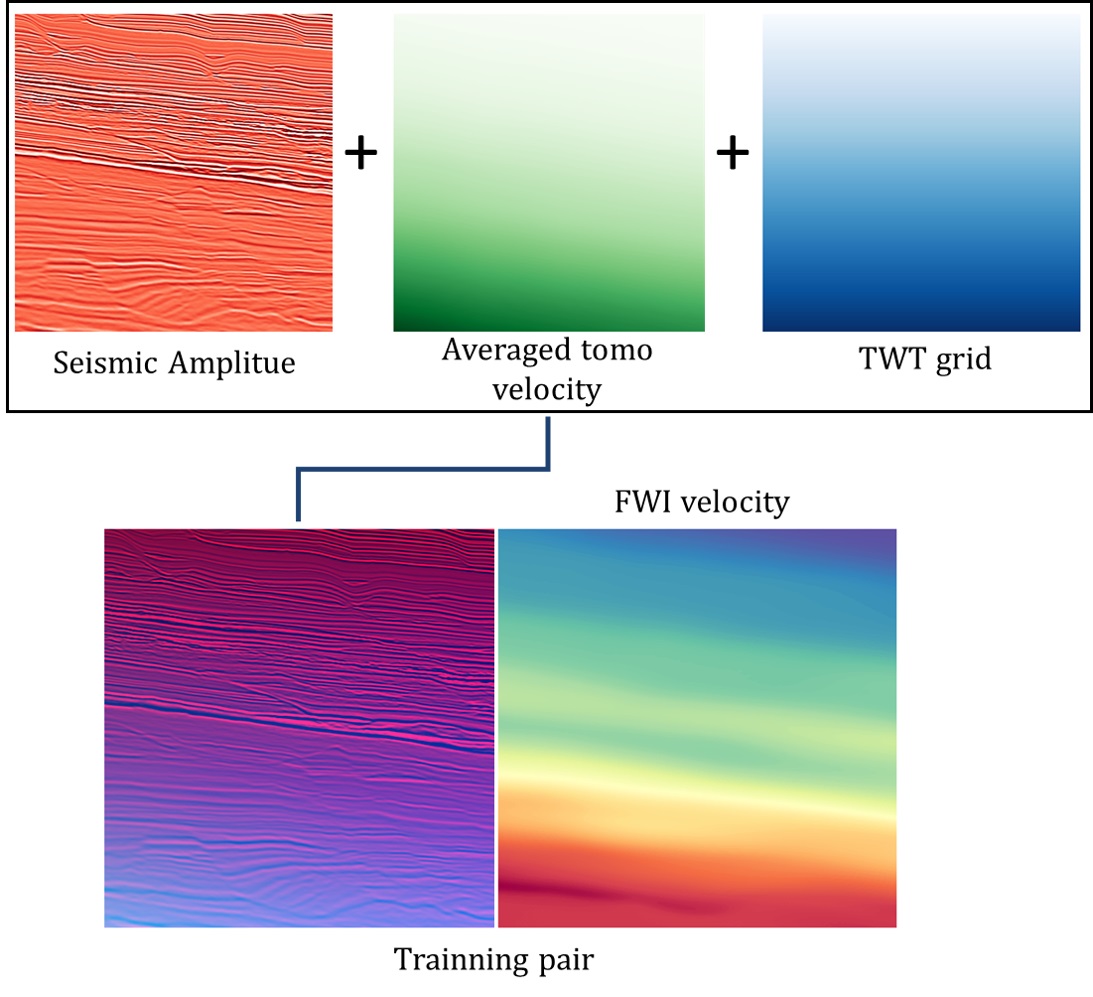}
  \caption{Three channels combined input and FWI velocity patches producing a training pair.}
 \label{fig:training_pair}
\end{figure}

\subsection{Network}

We used the pix2pix conditional GAN implemented in Pytorch with the UNet \cite{Ronneberger2015UNetCN} as a generator and PatchGAN\cite{isola2018imagetoimage} as the discriminator counterpart.

UNet is an encoder--decoder architecture with skip connections, and PatchGAN is a special type of classifier network subdividing the image into 70 × 70 patches and classifying each of them as real or fake (Fig. \ref{fig:GAN}). The GAN loss is the average loss of all patches. This strategy focuses on the high frequency of the image data, whereas the low-frequency content is captured by L1 loss.

\begin{figure}[hb]
 \centering
 \includegraphics[width=1.0\hsize]{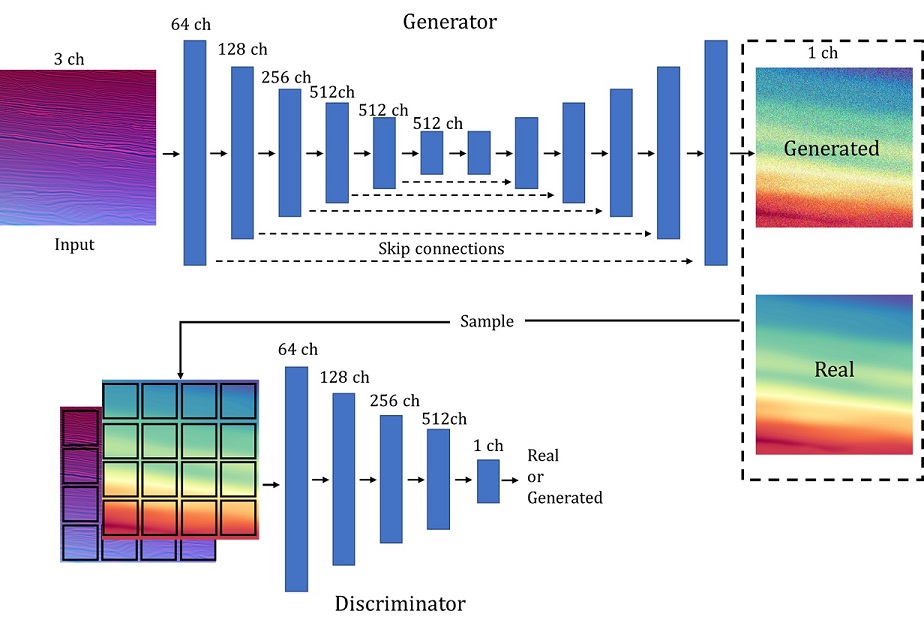}
  \caption{Schematic representation of the pix2pix network architecture.}
 \label{fig:GAN}
\end{figure}

\subsection{Experiments}

All experiments were performed using an HP Z8 workstation with a 24 GB memory Quadro P6000 GPU.
We began the experiment using only seismic images and associated FWI velocity as a training pair, but the generator produced artificial high frequency structure images with artifacts (Fig.  \ref{fig:oneChanel}). We conclude that seismic images alone did not provide enough information for the network to converge.

\begin{figure}[b]
 \centering
 \includegraphics[width=1.0\hsize]{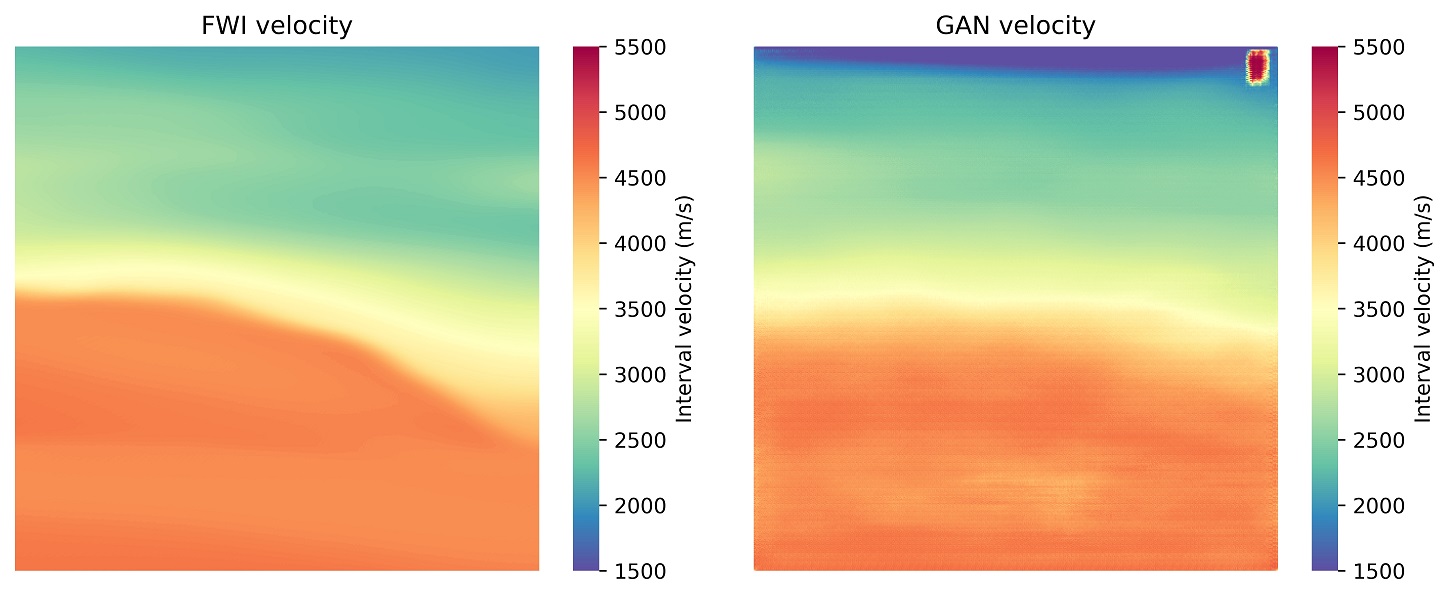}
  \caption{FWI velocity compared with one channel GAN velocity.}
 \label{fig:oneChanel}
\end{figure}

To obtain better results, we added two additional channels, the average tomographic velocity and the TWT grid, resulting in a three-channel input image (Fig. \ref{fig:training_pair}).

Although the tomographic interval velocity was inaccurate, it provided a reliable average velocity. Hence, we used average velocity rather than interval velocity. TWT information supported the network in better estimating spatial positions after we extracted the patches randomly, once there is a direct relation between velocity and overburden thickness.

Our assumption was that the tomographic average velocity and the TWT grid would function as boundary conditions, helping the network converge to the right interval velocity.
The training time for each experiment was approximately four hours. 

The generator and discriminator were trained using the same learning rate ($lr =\num{2e-5}$) which was established empirically, based on the visual quality of the training set results. We used the stochastic gradient descent method and the Adam solver with momentum parameters $\beta_1=0.5$ and $\beta_2=0.999$. The $L1$ multiplier $\lambda$ was set to $100$.

The results were evaluated based on three criteria: visual analysis, percent error (PE), and structural similarity index measure (SSIM) \cite{Wang04SSIM}.

Percent error is a common metric for quality control in seismic processing; it is defined as (\ref{eq:PE}).
\begin{align}
     \mathrm{PE}  =\frac{1}{N}\sum_{i=1}^{N}\frac{ \left |y(i)-{y}'(i)  \right |}{y(i)}*100.
     \label{eq:PE}
\end{align}

SSIM is a composite index (\ref{eq:SSMI_f}) measuring the similarity between two signals based on a comparison of their mean (\ref{eq:l}), variance (\ref{eq:c}), and correlation coefficient (\ref{eq:s}).

\begin{align}
  \mathrm{SSIM}(y,{y}') = f(l(y,{y}'),c(y,{y}'),s(y,{y}')),
  \label{eq:SSMI_f}
\end{align}

\begin{align}
    l(y,{y}') =\frac{2\mu_y\mu_{{y}'} + C_1}{\mu_y^2 + \mu_{{y}'}^2+ C_1},  \label{eq:l}
\end{align}

\begin{align}
    c(y,{y}') =\frac{2\sigma_y\sigma_{{y}'} + C_2}{\sigma_y^2 + \sigma_{{y}'}^2+ C_2},  \label{eq:c}
\end{align}

\begin{align}
    s(y,{y}') =\frac{\sigma_{y{y}'} + C_3}{\sigma_y \sigma_{{y}'}+ C_3},  \label{eq:s}
\end{align}

where $C_1$, $C_2$, and $C_3$ are constants to avoid division by zero. Setting $C_3 = \frac{C_2}{2}$, we obtain the SSIM equation (\ref{eq:SSMI}).

\begin{align}
  \mathrm{SSIM}(y,{y}') = \frac{(2\mu_y\mu_{{y}'} + C_1) + (2 \sigma _{y{y}'} + C_2)} 
    {(\mu_y^2 + \mu_{{y}'}^2+C_1) (\sigma_y^2 + \sigma_{{y}'}^2+C_2)}.
  \label{eq:SSMI}
\end{align}

\begin{figure*}[tbh]
 \centering
 \includegraphics[width=1.0\hsize]{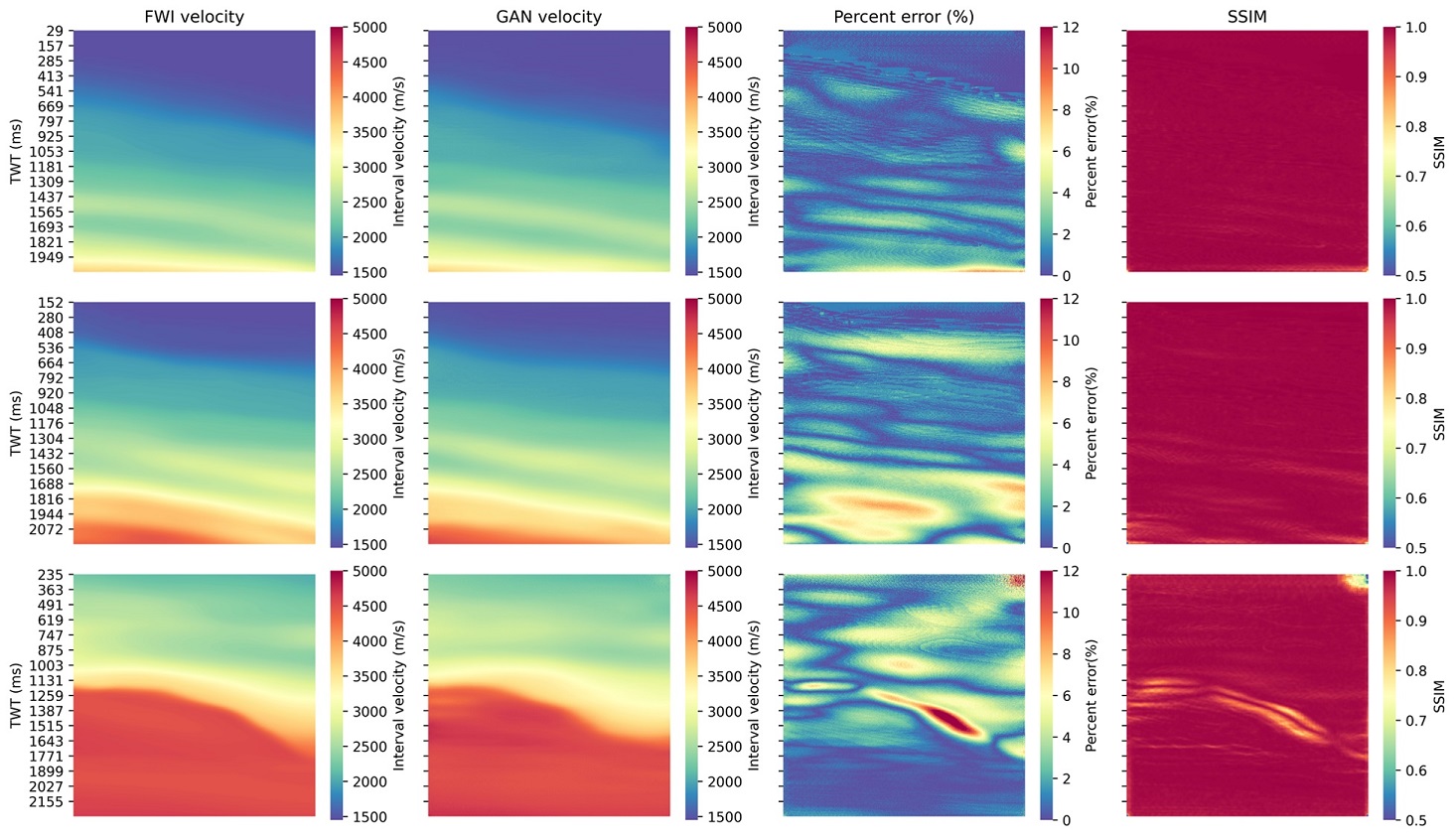}
  \caption{FWI velocity, GAN velocity, percent error, and SSIM map over three representative patches (shallow, intermediate, and deep regions from top to bottom, respectively).}
 \label{fig:results}
\end{figure*}
\section{Results}

Our pix2pix GAN approach produced an accurate velocity model using three channels (post-stack seismic image, average tomographic velocity, and two-way time) as input (Fig.~\ref{fig:results}).
The chosen metrics show excellent results on the test set. The average percent error was  $1.5\%$, average SSIM was 0.994, and the generated velocity was visually identical to the FWI velocity, without artifacts, verifying that the geological structures were incorporated into the GAN velocity model. 
The two distributions are very close, confirming the metric results (Fig. \ref{fig:vel_distrib}).

The criterion to stop training was set by visual analysis of the training set results and later confirmed using the percent error and the SSIM results based on the test set. Both metrics stabilized between 60 and 80 epochs (Fig. \ref{fig:PE} and Fig. \ref{fig:SSIM}).

\begin{figure}[bth]
 \centering
 \includegraphics[width=1\hsize]{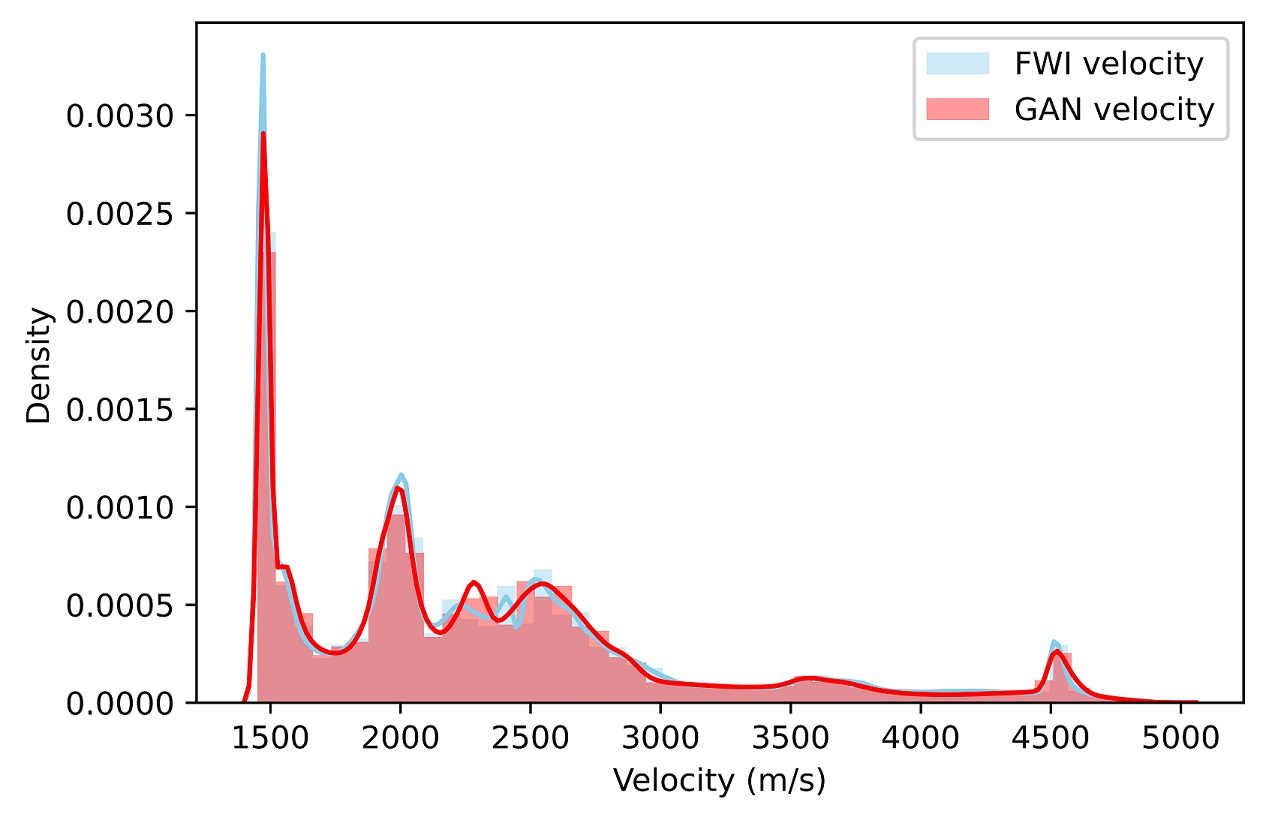}
  \caption{FWI and GAN velocity distributions of test set.}
 \label{fig:vel_distrib}
\end{figure}

Unexpectedly, the one-channel model experiment showed satisfactory SSIM and PE results when compared with the three-channel model, although the visual analysis showed very poor results. 

Shallower regions presented better results than deeper regions on all metrics (Fig.  \ref{fig:results}).

\section{Discussion}

Previous works \cite{Puzyrev2019SeismicIW} demonstrated the feasibility of using CNNs for one-dimensional (1D) synthetic seismic inversion, and \cite{Wu2018InversionNetAA} used CNNs to perform seismic inversion with synthetic pre-stack seismic data. Conditional unpaired CycleGANs\cite{Mosser2018RapidSD} were employed to perform forward and inverse seismic modeling with post-stack seismic data and velocity models, while also using synthetic data with a single input channel (post-stack seismic image). They achieved only qualitatively moderate results.
GANs were also used to generate prior models to constrain solutions in the traditional FWI algorithm\cite{Mosser2020}.

In this study, we apply a conditional pix2pix GAN to generate highly accurate velocity models using real post-stack seismic data with multiple inputs.
\begin{figure}[t]
 \centering
 \includegraphics[width=1\hsize]{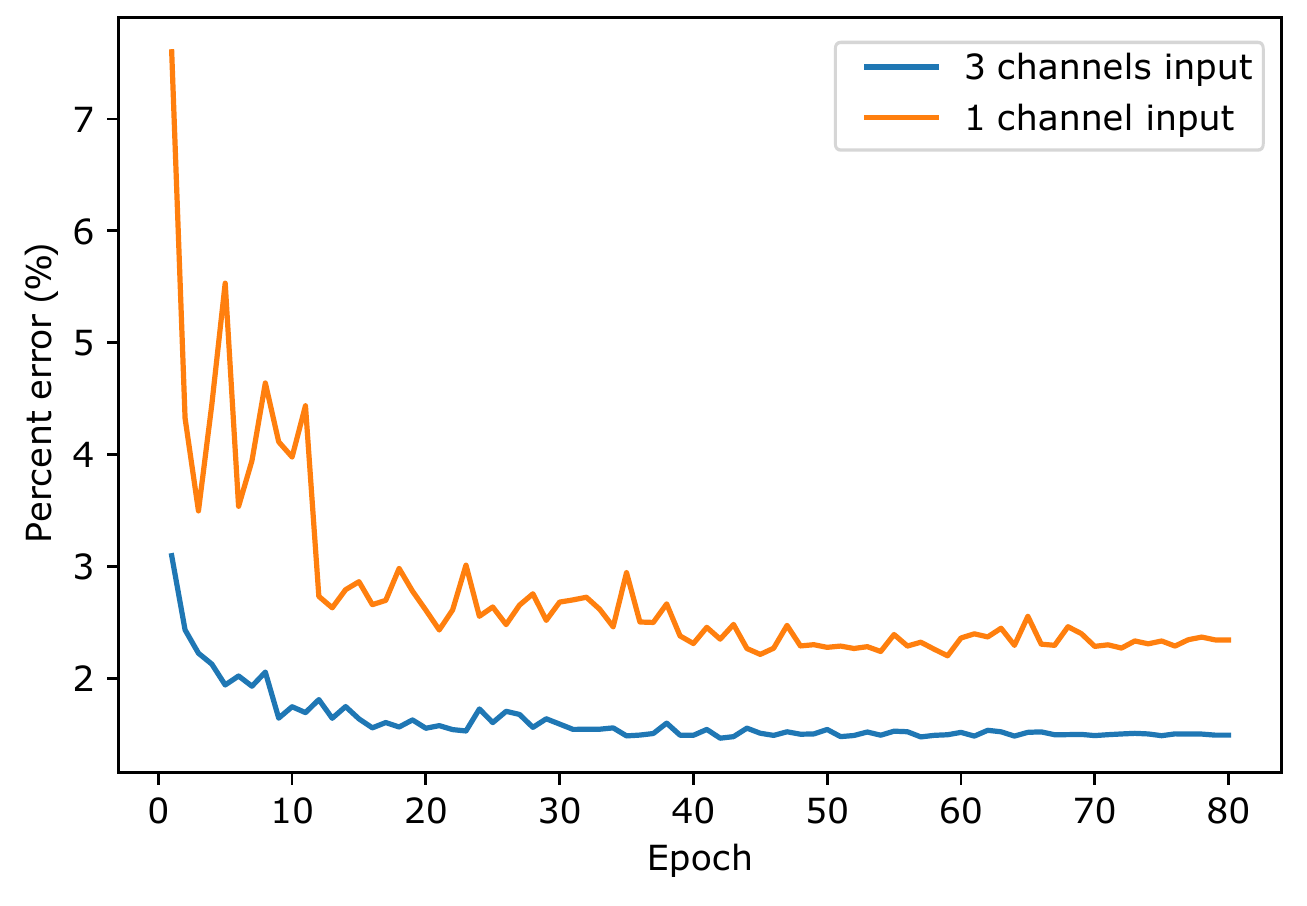}
  \caption{Percent error versus epochs using test set.}
 \label{fig:PE}
\end{figure}

\begin{figure}[h]
 \centering
 \includegraphics[width=1\hsize]{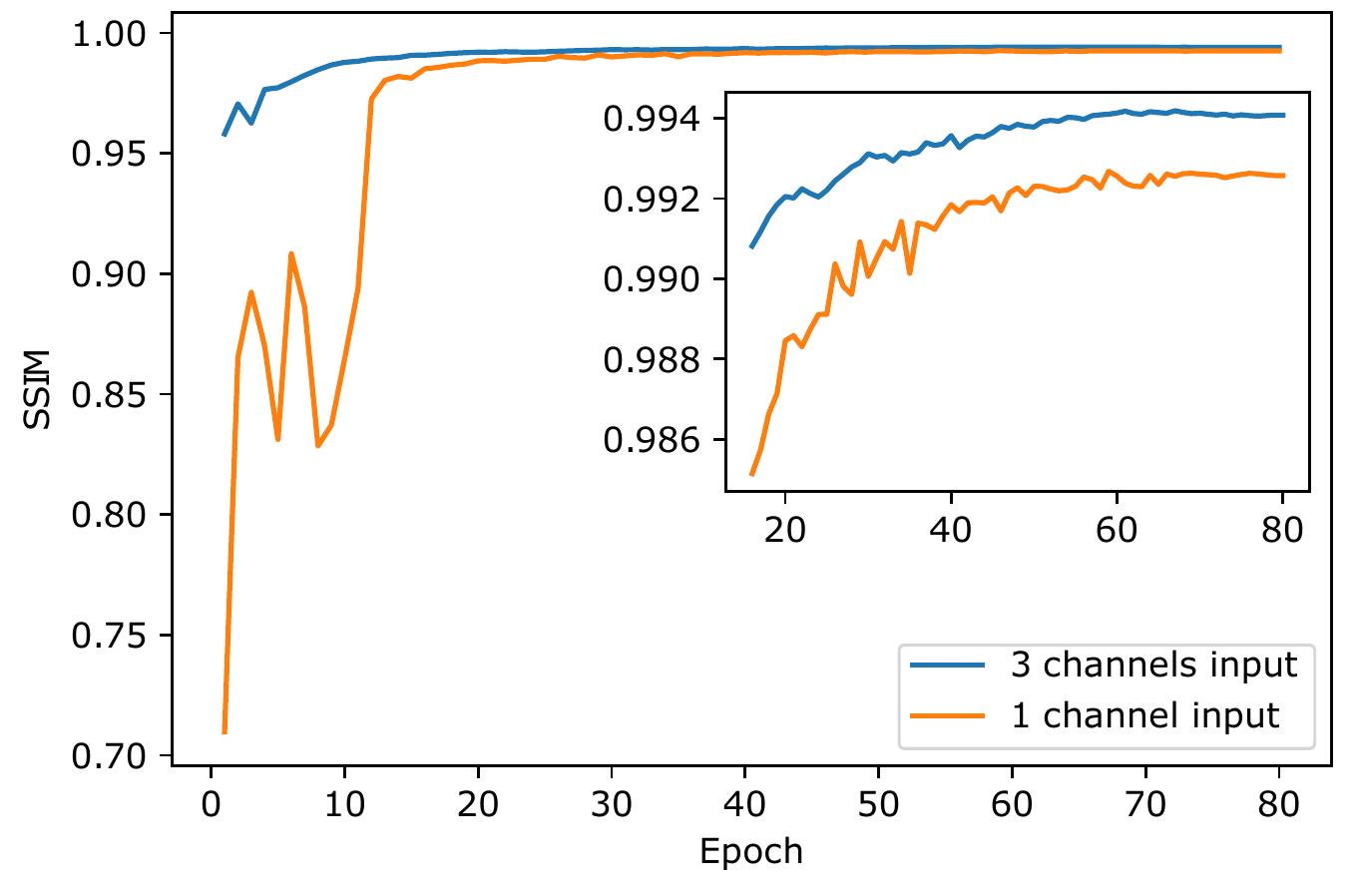}
  \caption{SSIM versus epochs using test set.}
 \label{fig:SSIM}
\end{figure}
Neural networks are understood to perform better at interpolation than extrapolation tasks \cite{Barnard1992ExtrapolationAI}, \cite{Xu2020HowNN}; therefore, it is challenging for a cGAN to extrapolate to lower frequencies using only the post-stack seismic image as input. However, it becomes easier to interpolate the missing frequencies (Fig.~\ref{fig:FWIgap}) if we combine lower frequency data as input (in this case, average tomographic velocity and TWT grid). This explains why adding more information helped to improve the results.

Providing a TWT grid and the average tomographic velocity enabled a depth estimation for each sample, which is useful information for estimating the interval velocity as it increases with depth.

Ordinary physics-based FWI provides better results in shallower regions, where the acquisition cables have sufficient length to capture diving waves \cite{Zhou2015FullWI} and worse results in deeper regions. As expected, the pix2pix GAN captured this inaccuracy and also produced worse results in deeper regions (Fig. \ref{fig:results}). 

The $L1$ regularization loss acts as a global metric, minimizing the average loss of all pixels, whereas the GAN loss, provided by the PatchGAN discriminator, focuses on high-frequency structures. We suspect that the reason the experiment using one channel presented good PE and SSIM values was that the generator produced a velocity field that honors the low frequency associated to average velocity (good PE) and the high frequency variations (good SSIM), however producing poorly accurate visual results. This result reveals the difficulty of finding reliable metrics for evaluating GANs \cite{Borji2019ProsAC}.

The major threat to the method is the usage of poor-quality training data. Despite the robustness of FWI velocity modeling, examples where FWI could not converge to an acceptable velocity model because of seismic acquisition limitations and low signal-to-noise ratio are common\cite{yao2020review}. If this data were used to train the model, the pix2pix cGAN will also produce a poor-quality velocity model.

GANs are normally difficult to train, they are prone to reduce the output variability or a complete mode collapse \cite{salimans2016improved}. Different architectures such as Variational Autoencoders and Autoencoder with perceptual loss need to be tested in order to avoid such problems.

\section{Conclusions}
We have presented a method, based on a pix2pix conditional GAN with multiple input, that can be used to surrogate FWI velocity modeling. The three inputs (post-stack seismic image, average tomographic velocity, and TWT grid) were vital, as the experiment using only one channel (post-stack seismic image) failed. 

Further development of new network architectures focused on seismic data particularities and considering  three or more spatial dimensions   may generate an improvement in the quality of our results.

The test data presented excellent results based on PE, SSIM, and visual analysis. The main geological structures and lateral velocity changes were captured by the resulting velocity model.

Once we train the GAN, it was really fast the generation of a FWI interval velocity model surrogate, almost real time. Therefore, our proposed method can replace the FWI modeling or providing an initial velocity model that will need fewer iterations to converge during traditional FWI, similar strategy adopted by\cite{senior2020improved}

A fair comparison between the FWI and our cGAN surrogate model should consider the implementation of the FWI algorithm in parallel, using GPU hardware. Although some papers have proposed GPU implementations of the FWI algorithm\cite{Wang2011CUDAbasedAO}, \cite{Mao2012MultiscaleFW} and \cite{Fang2020ElasticFI}, they only used 2D synthetic data, mainly due to GPU memory constraints. At present, FWI algorithm processing on GPU hardware has yet to become industrially feasible.

The proposed method, when deployed, has the potential to speed up geophysical reservoir characterization processes, saving time and reducing computational resource requirements.

\bibliographystyle{IEEEtran}
\bibliography{IEEEexample}

\begin{thebibliography}{10}
\providecommand{\url}[1]{#1}
\csname url@samestyle\endcsname
\providecommand{\newblock}{\relax}
\providecommand{\bibinfo}[2]{#2}
\providecommand{\BIBentrySTDinterwordspacing}{\spaceskip=0pt\relax}
\providecommand{\BIBentryALTinterwordstretchfactor}{4}
\providecommand{\BIBentryALTinterwordspacing}{\spaceskip=\fontdimen2\font plus
\BIBentryALTinterwordstretchfactor\fontdimen3\font minus
  \fontdimen4\font\relax}
\providecommand{\BIBforeignlanguage}[2]{{%
\expandafter\ifx\csname l@#1\endcsname\relax
\typeout{** WARNING: IEEEtran.bst: No hyphenation pattern has been}%
\typeout{** loaded for the language `#1'. Using the pattern for}%
\typeout{** the default language instead.}%
\else
\language=\csname l@#1\endcsname
\fi
#2}}
\providecommand{\BIBdecl}{\relax}
\BIBdecl

\bibitem{Gazdag_Sguazzero}
J.~{Gazdag} and P.~{Sguazzero}, ``Migration of seismic data,''
  \emph{Proceedings of the IEEE}, vol.~72, no.~10, pp. 1302--1315, 1984.

\bibitem{Yilmaz}
\BIBentryALTinterwordspacing
O.~Yilmaz, \emph{{Seismic Data Analysis: Processing, Inversion, and
  Interpretation of Seismic Data}}.\hskip 1em plus 0.5em minus 0.4em\relax
  Society of Exploration Geophysicists, 01 2001. [Online]. Available:
  \url{https://doi.org/10.1190/1.9781560801580}
\BIBentrySTDinterwordspacing

\bibitem{schultz1998seismic}
P.~Schultz, \emph{The seismic velocity model as an interpretation asset}.\hskip
  1em plus 0.5em minus 0.4em\relax Society of Exploration Geophysicists, 1998.

\bibitem{Liu2009StackingSD}
G.~Liu, S.~Fomel, L.~Jin, and X.~Chen, ``Stacking seismic data using local
  correlation,'' \emph{Geophysics}, vol.~74, 2009.

\bibitem{Kapoor2012FullWI}
S.~Kapoor, D.~Vigh, H.~Li, and D.~Derharoutian, ``Full waveform inversion for
  detailed velocity model building,'' in \emph{74th EAGE Conference and
  Exhibition incorporating EUROPEC 2012}.\hskip 1em plus 0.5em minus
  0.4em\relax European Association of Geoscientists \& Engineers, 2012, pp.
  cp--293.

\bibitem{isola2018imagetoimage}
P.~Isola, J.-Y. Zhu, T.~Zhou, and A.~A. Efros, ``Image-to-image translation
  with conditional adversarial networks,'' 2018.

\bibitem{Jones2010TutorialVE}
I.~Jones, ``Tutorial: Velocity estimation via ray-based tomography,''
  \emph{First Break}, vol.~28, 2010.

\bibitem{lailly1983sequence}
P.~Lailly, ``The seismic inverse problem as a sequence of before stack
  inversions,'' in \emph{Conference on Inverse Scattering--Theory and
  Application}, vol.~11.\hskip 1em plus 0.5em minus 0.4em\relax Siam, 1983, p.
  206.

\bibitem{Tarantola1984InversionOS}
A.~Tarantola, ``Inversion of seismic reflection data in the acoustic
  approximation,'' \emph{Geophysics}, vol.~49, pp. 1259--1266, 1984.

\bibitem{Madagascar}
{Madagascar Development Team}, \emph{Madagascar Software, Version~1.4},
  http://www.ahay.org/, 2012.

\bibitem{virieux2017introduction}
J.~Virieux, A.~Asnaashari, R.~Brossier, L.~M{\'e}tivier, A.~Ribodetti, and
  W.~Zhou, ``An introduction to full waveform inversion,'' in
  \emph{Encyclopedia of exploration geophysics}.\hskip 1em plus 0.5em minus
  0.4em\relax Society of Exploration Geophysicists, 2017, pp. R1--1.

\bibitem{Geng2020AutomatedDO}
Z.~Geng and Y.~Wang, ``Automated design of a convolutional neural network with
  multi-scale filters for cost-efficient seismic data classification,''
  \emph{Nature Communications}, vol.~11, 2020.

\bibitem{Xiong2018SeismicFD}
W.~Xiong, X.~Ji, Y.~Ma, Y.~Wang, N.~M. AlBinHassan, M.~N. Ali, and Y.~Luo,
  ``Seismic fault detection with convolutional neural network,''
  \emph{Geophysics}, vol.~83, 2018.

\bibitem{Shi2019SaltSegA3}
Y.~Shi, X.~Wu, and S.~Fomel, ``Saltseg: Automatic 3d salt segmentation using a
  deep convolutional neural network,'' \emph{Interpretation}, vol.~7, 2019.

\bibitem{Wu2019FaultSeg3DUS}
X.~Wu, L.~Liang, Y.~Shi, and S.~Fomel, ``Faultseg3d: Using synthetic data sets
  to train an end-to-end convolutional neural network for 3d seismic fault
  segmentation,'' \emph{Geophysics}, vol.~84, 2019.

\bibitem{Guazzelli2020Efficient3S}
A.~B. Guazzelli, M.~Roisenberg, and B.~Rodrigues, ``Efficient 3d semantic
  segmentation of seismic images using orthogonal planes 2d convolutional
  neural networks,'' \emph{2020 International Joint Conference on Neural
  Networks (IJCNN)}, pp. 1--8, 2020.

\bibitem{Ren2020APN}
Y.~Ren, X.~Xu, S.~Yang, L.~Nie, and Y.~Chen, ``A physics-based neural-network
  way to perform seismic full waveform inversion,'' \emph{IEEE Access}, vol.~8,
  pp. 112\,266--112\,277, 2020.

\bibitem{Moseley2020SolvingTW}
B.~Moseley, A.~Markham, and T.~Nissen-Meyer, ``Solving the wave equation with
  physics-informed deep learning,'' \emph{arXiv: Computational Physics}, 2020.

\bibitem{Zhang2019PhysicsguidedCN}
R.~Zhang, Y.~Liu, and H.~Sun, ``Physics-guided convolutional neural network
  (phycnn) for data-driven seismic response modeling,'' \emph{ArXiv}, vol.
  abs/1909.08118, 2019.

\bibitem{Claerbout1985ImagingTE}
J.~F. Claerbout, \emph{Imaging the earth's interior}.\hskip 1em plus 0.5em
  minus 0.4em\relax Blackwell scientific publications Oxford, 1985, vol.~1.

\bibitem{Goodfellow2014GenerativeAN}
I.~J. Goodfellow, J.~Pouget-Abadie, M.~Mirza, B.~Xu, D.~Warde-Farley, S.~Ozair,
  A.~C. Courville, and Y.~Bengio, ``Generative adversarial nets,'' in
  \emph{NIPS}, 2014.

\bibitem{Pathak2016ContextEF}
D.~Pathak, P.~Kr{\"a}henb{\"u}hl, J.~Donahue, T.~Darrell, and A.~A. Efros,
  ``Context encoders: Feature learning by inpainting,'' \emph{2016 IEEE
  Conference on Computer Vision and Pattern Recognition (CVPR)}, pp.
  2536--2544, 2016.

\bibitem{segyiodocs}
\BIBentryALTinterwordspacing
Equinor. {segyio documentation}. [Online]. Available:
  \url{https://segyio.readthedocs.io/en/latest/index.html}
\BIBentrySTDinterwordspacing

\bibitem{Ronneberger2015UNetCN}
O.~Ronneberger, P.~Fischer, and T.~Brox, ``U-net: Convolutional networks for
  biomedical image segmentation,'' \emph{ArXiv}, vol. abs/1505.04597, 2015.

\bibitem{Wang04SSIM}
Z.~Wang, A.~C. Bovik, H.~R. Sheikh, S.~Member, E.~P. Simoncelli, and S.~Member,
  ``Image quality assessment: From error visibility to structural similarity,''
  \emph{IEEE Transactions on Image Processing}, vol.~13, pp. 600--612, 2004.

\bibitem{Puzyrev2019SeismicIW}
V.~Puzyrev, A.~Egorov, A.~Pirogova, C.~Elders, and C.~Otto, ``Seismic inversion
  with deep neural networks: A feasibility analysis,'' in \emph{81st EAGE
  Conference and Exhibition 2019}, vol. 2019, no.~1.\hskip 1em plus 0.5em minus
  0.4em\relax European Association of Geoscientists \& Engineers, 2019, pp.
  1--5.

\bibitem{Wu2018InversionNetAA}
Y.~Wu, Y.~Lin, and Z.~Zhou, ``Inversionnet: Accurate and efficient seismic
  waveform inversion with convolutional neural networks,'' \emph{Seg Technical
  Program Expanded Abstracts}, 2018.

\bibitem{Mosser2018RapidSD}
L.~Mosser, W.~Kimman, J.~Dramsch, S.~Purves, A.~D.~L. Fuente, and G.~Ganssle,
  ``Rapid seismic domain transfer: Seismic velocity inversion and modeling
  using deep generative neural networks,'' \emph{ArXiv}, vol. abs/1805.08826,
  2018.

\bibitem{Mosser2020}
\BIBentryALTinterwordspacing
L.~Mosser, O.~Dubrule, and M.~J. Blunt, ``{Stochastic Seismic Waveform
  Inversion Using Generative Adversarial Networks as a Geological Prior},''
  \emph{Mathematical Geosciences}, vol.~52, no.~1, pp. 53--79, jan 2020.
  [Online]. Available: \url{http://arxiv.org/abs/1806.03720
  http://link.springer.com/10.1007/s11004-019-09832-6}
\BIBentrySTDinterwordspacing

\bibitem{Barnard1992ExtrapolationAI}
E.~Barnard and L.~Wessels, ``Extrapolation and interpolation in neural network
  classifiers,'' \emph{IEEE Control Systems}, vol.~12, pp. 50--53, 1992.

\bibitem{Xu2020HowNN}
K.~Xu, J.~Li, M.~Zhang, S.~Du, K.~Kawarabayashi, and S.~Jegelka, ``How neural
  networks extrapolate: From feedforward to graph neural networks,''
  \emph{ArXiv}, vol. abs/2009.11848, 2020.

\bibitem{Zhou2015FullWI}
W.~Zhou, R.~Brossier, S.~Operto, and J.~Virieux, ``Full waveform inversion of
  diving \& reflected waves for velocity model building with impedance
  inversion based on scale separation,'' \emph{Geophysical Journal
  International}, vol. 202, pp. 1535--1554, 2015.

\bibitem{Borji2019ProsAC}
A.~Borji, ``Pros and cons of gan evaluation measures,'' \emph{Comput. Vis.
  Image Underst.}, vol. 179, pp. 41--65, 2019.

\bibitem{yao2020review}
G.~Yao, D.~Wu, and S.-X. Wang, ``A review on reflection-waveform inversion,''
  \emph{Petroleum Science}, vol.~17, no.~2, pp. 334--351, 2020.

\bibitem{salimans2016improved}
T.~Salimans, I.~Goodfellow, W.~Zaremba, V.~Cheung, A.~Radford, and X.~Chen,
  ``Improved techniques for training gans,'' \emph{arXiv preprint
  arXiv:1606.03498}, 2016.

\bibitem{senior2020improved}
A.~W. Senior, R.~Evans, J.~Jumper, J.~Kirkpatrick, L.~Sifre, T.~Green, C.~Qin,
  A.~{\v{Z}}{\'\i}dek, A.~W. Nelson, A.~Bridgland \emph{et~al.}, ``Improved
  protein structure prediction using potentials from deep learning,''
  \emph{Nature}, vol. 577, no. 7792, pp. 706--710, 2020.

\bibitem{Wang2011CUDAbasedAO}
B.~li~Wang, J.~Gao, H.~Zhang, and W.~Zhao, ``Cuda-based acceleration of full
  waveform inversion on gpu,'' \emph{Seg Technical Program Expanded Abstracts},
  2011.

\bibitem{Mao2012MultiscaleFW}
J.~Mao, R.~shan Wu, and B.~Wang, ``Multiscale full waveform inversion using
  gpu,'' \emph{Seg Technical Program Expanded Abstracts}, 2012.

\bibitem{Fang2020ElasticFI}
J.~Fang, H.~Chen, H.~Zhou, Y.~Rao, P.~Sun, and J.~Zhang, ``Elastic
  full-waveform inversion based on gpu accelerated temporal fourth-order
  finite-difference approximation,'' \emph{Comput. Geosci.}, vol. 135, p.
  104381, 2020.

\end{thebibliography}

\end{document}